\documentclass{article}

\usepackage[dblblindworkshop, final]{neurips_2026}
\workshoptitle{NewInML}
\makeatletter
\renewcommand{\@noticestring}{%
  NewInML Workshop at the 40th Conference on Neural Information Processing Systems (NeurIPS 2026), Paris, France, 2026.
}
\makeatother
\usepackage{amsmath}

\usepackage[utf8]{inputenc} 
\usepackage[T1]{fontenc}    
\usepackage[hidelinks]{hyperref}       
\usepackage{url}            
\usepackage{booktabs}       
\usepackage{amsfonts}       
\usepackage{nicefrac}       
\usepackage{microtype}      
\usepackage{xcolor}         
\usepackage{graphicx}

\newcommand{\PAG}{\mathrm{PAG}}
\newcommand{\RH}{R_{H}}
\newcommand{\RL}{R_{L}}
\newcommand{\Pnov}{P_{\mathrm{nov}}}
\newcommand{\Padv}{P_{\mathrm{adv}}}

\title{Difference-in-Differences on a Censored Rating Scale Can Manufacture an Effect: Evidence from a Pre-Registered LLM-Judge Audit}

\author{%
  Shuyi Fan\thanks{Equal contribution.} \\
  Columbia University \\
  \texttt{ff2483@tc.columbia.edu} \\
  \And
  Boyuan Deng\footnotemark[1] \\
  Johns Hopkins University \\
  \texttt{bdeng8@jh.edu} \\
  \AND
  Mengyu Xu\footnotemark[1] \\
  The University of Chicago \\
  \texttt{mxu09@uchicago.edu} \\
  \And
  Xinhong Xie \\
  The Pennsylvania State University \\
  \texttt{xjx5116@psu.edu} \\
  \And
  Chenyang Li \\
  Johns Hopkins University \\
  \texttt{chenyangli2020@u.northwestern.edu} \\
  \And
  Hongyang Zhang\thanks{Corresponding author.} \\
  The Hong Kong Polytechnic University \\
  \texttt{hong-yang.zhang@connect.polyu.hk} \\
}

\begin{document}

\maketitle

\begin{abstract}
    LLM-judge audits assess bias by comparing ratings across matched conditions. Difference-in-differences designs take the rating difference between two candidate responses to each item and compare it across a manipulated attribute. Building on prior work on nonlinear difference-in-differences, we show in closed form that when ratings are bounded, a severity shift common to both responses produces an observed interaction whenever the scale bounds attenuate it unequally. We then report a pre-registered audit of 990 calls to a frozen pedagogy judge. The registered primary analysis did not detect an effect of a stated novice or advanced profile on the judge's preference for a high-scaffolding over a low-scaffolding response. The only nominally significant secondary interaction was on the productive-struggle sub-score ($+0.378$ points, $p = 0.002$). A post-hoc construction in which the profile shifts both responses' latent ratings equally reproduced 79 to 85\% of this interaction using the observed severity shift and the scale floor alone. The observed interaction therefore does not identify differential preference on the latent scale. Code is available at \url{https://github.com/bydeng01/ability-levels-audit}.
\end{abstract}

\section{Introduction}
\label{sec:intro}

Large language models now grade other language models. Scores emitted by an LLM judge steer leaderboards and benchmark rankings \citep{zheng2023judging,gu2026survey}, and in education LLM judges already score tutoring systems on rubrics of teaching quality \citep{maurya2025unifying,fan2026rethinkingllmjudgedhelpfulnesspedagogy}. Whether such judges deserve that role is itself a measurement question, and the audits that answer it share a design: hold the item fixed, vary an attribute of its presentation or of the identity attached to it, and read the bias off the contrast between matched conditions \citep{wang2024large,howell2025prestige}. A stronger design differences twice: a within-item contrast between two candidate responses is differenced again across the manipulated attribute, so that anything shifting a response's score equally in every condition cancels. The resulting endpoint is a difference in differences on a bounded rating scale.

The design's interpretation also rests on an assumption: when an effect survives the double difference, the bounded scale is treated as conservative, since clipping at the scale ends (censoring) can attenuate a real effect but not create one. For a single contrast that is true, because clipping never widens the gap between two values. Our own pre-registration asserted it for the double difference, and there it is false. A common severity shift makes the judge harsher or less strict toward both responses by the same latent amount. Such a shift can produce a nonzero interaction when the scale bounds attenuate it differently for the two responses (\S\ref{sec:mech}). Econometrics has long separated the latent interaction from the difference in differences of observed outcomes, yet LLM-judge audits report interaction effects from short bounded rubrics without identifiability checks (e.g., the factorial interactions of \citet{maltbie2026intersectional}, read off a 1--10 rubric with most ratings within two points of the floor). This risk grows with the quality of the materials: sharper contrasts push the two responses toward opposite ends of the scale, where a common shift is censored most unequally.

This paper demonstrates the failure in the data of a pre-registered audit. The study put a validity question to the frozen pedagogy judge of \citet{fan2026rethinkingllmjudgedhelpfulnesspedagogy}, which scores candidate tutor turns on a four-principle pedagogy rubric: does its preference for high-scaffolding responses track the competence a learner demonstrates in the dialogue, or does a stated learner profile move it? Each of 55 frozen rating contexts pairs a high-scaffolding with a low-scaffolding candidate response, the two poles of that context. The judge rates both in three arms, under no profile, a stated novice profile, and a stated advanced profile. The registered primary endpoint, the Profile Anchoring Gap (PAG), is the change in the judge's scaffolding preference between the two profiles on the weak stratum, the stimuli where the learner visibly struggles.

The registered endpoint is null: against a baseline preference of $+2.6$ to $+3.2$ points of the five-point scale, the profile moves the preference by $+0.085$ (95\% bootstrap interval $[-0.167, +0.353]$, exact signed-rank $p = 0.684$, \S\ref{sec:primary}). The profile did, however, affect absolute ratings. Both poles received lower ratings under the advanced profile, consistent with a common severity shift. The scale can convert such a shift into the audit's one nominally significant interaction: a $+0.378$ gap on the productive-struggle sub-score ($p = 0.002$). A construction in which the profile shifts both poles' latent ratings equally reproduces 79 to 85\% of that gap from the observed high-pole shift and the scale floor alone (Figure~\ref{fig:two}c). On 17 of the 30 weak-stratum stimuli the productive-struggle low pole is pinned at the floor in every arm, and there the difference in differences is algebraically a one-pole contrast.

This paper makes three contributions. First, an identifiability analysis of within-item difference-in-differences endpoints on bounded rating scales. The statistical result is known for limited-dependent-variable models \citep{puhani2012treatment}. We show that in judge audits the endpoint approaches a one-pole contrast as the two candidate responses are rated further apart. Second, a pre-registered audit of profile influence on a frozen pedagogy judge. The primary endpoint is null. The secondaries show that censoring can manufacture an effect, contrary to the audit's own registered assumption. Third, an identifiability check that follows from the same algebra, computable from ratings an audit already has.

\section{Related work}
\label{sec:related}

\paragraph{LLM judges and their audits.}
Judge-based evaluation entered common use with MT-Bench and Chatbot Arena, which reported high judge--human agreement alongside position, verbosity, and self-enhancement biases \citep{zheng2023judging}. Later audits extended the catalog: order effects that flip verdicts \citep{wang2024large}, self-preference in proportion to self-recognition \citep{panickssery2024llm}, and diverging human and model bias profiles under controlled perturbation \citep{chen2024humans}. Surveys collect both the biases and the audit designs that detect them \citep{gu2026survey}. We examine whether difference-in-differences contrasts on bounded rating scales identify latent differential preference.

\paragraph{Sensitivity to stated attributes.}
A second line varies who the text is about or from. Sociodemographic prompting shifts model predictions on subjective tasks, though inconsistently across models and datasets \citep{beck2024sensitivity}. Persona variables explain a small fraction of human annotation variance \citep{hu2024quantifying}. Stated institutional prestige moves simulated peer-review outcomes for otherwise identical manuscripts \citep{howell2025prestige}. These findings motivate our manipulated factor, a stated learner profile, and suggest that its effect is likely to be small.

\paragraph{Pedagogical evaluation of tutoring systems.}
The rubric principles at stake descend from scaffolding \citep{wood1976role} and productive failure \citep{kapur2008productive}, operationalized for LLM tutors in dialogue benchmarks, error-remediation corpora, and evaluation taxonomies \citep{tack2022ai,macina2023mathdial,daheim2024stepwise,maurya2025unifying,learnlm2024}. The instrument we audit is the frozen pedagogy judge of \citet{fan2026rethinkingllmjudgedhelpfulnesspedagogy}, which found generic helpfulness ratings judge-dependent while rubric distinctions were stable. Its corpus supplies our stimuli.

\paragraph{Interactions on bounded and ordinal scales.}
Outside NLP, the gap between observed and latent interactions is an old problem. Censored and ordinal reporting make the observed score a nonlinear transform of a latent quantity \citep{tobin1958estimation,mckelvey1975statistical}. In such models the interaction of observed conditional means and the interaction coefficient are different quantities \citep{ai2003interaction}, and the observed cross difference is not the treatment effect in nonlinear difference-in-differences models \citep{puhani2012treatment}. Identification that survives monotone rescaling of the outcome requires the stronger assumptions of \citet{athey2006identification}, and for ordinal outcomes the estimand is not even well defined without latent-scale assumptions \citep{yamauchi2020difference}. Treating ordinal ratings as interval data misleads even for ratings well away from the scale endpoints \citep{liddell2018analyzing}, and lacks justification in NLG \citep{amidei2019use}. \citet{liddell2018analyzing} further show that metric models of ordinal ratings produce false-alarm, missed and inverted \emph{interactions} in factorial designs. That a scale can manufacture an interaction is known more generally \citep{rohrer2021precise}. Recent work on judge scale design tunes granularity for human agreement \citep{li2026grading} but does not treat the bounds as an identification problem. We carry the problem into LLM-judge audits in its within-item, two-pole form.

\paragraph{Pre-registration in ML evaluation.}
Data-dependent analysis invalidates a reported $p$-value even when only one test is ever run \citep{gelman2016statistical}, and pre-registration is advocated for NLP \citep{van2021preregistering} and for predictive modeling \citep{hofman2023pre}. This study followed the registered analysis, but one of the pre-registration's interpretive assumptions proved false.

\section{Study design and materials}
\label{sec:setup}

\paragraph{Question and manipulation.}
The manipulation applies only to the judge's input: no new tutoring dialogues were generated. Each stimulus is a frozen rating context ending in a student turn, paired with two candidate next tutor turns. The high-scaffolding pole $\RH$ addresses where the student is and leaves the next reasoning step to them. The low-scaffolding pole $\RL$ performs that step for them. High scaffolding refers to contingent support, not simply more help. Each (stimulus, response) is rated in three arms: $D$ (dialogue and response only), $\Pnov$ (a stated novice learner profile prepended), and $\Padv$ (a stated advanced profile prepended). We call the manipulated variable a stated learner profile rather than an ability label, because the two texts are not a minimal contrast on ability: each also states a help-seeking preference, which is the construct the assistance-calibration sub-score measures directly. We therefore report the endpoint as profile influence.

\paragraph{Materials.}
We selected 55 stimuli from the 2,219 judged tutor turns identified in the released logs of \citet{fan2026rethinkingllmjudgedhelpfulnesspedagogy}. The stimuli span six acid-mixture word problems and 23 source tutoring runs, with weak and strong competence strata defined by blinded labels (Table~\ref{tab:materials}). The corpus has two kinds of tutor on the same base models. A pedagogy-tuned tutor estimates the learner's state before each reply. A conversational tutor replies in a single model call. One pole of each pair is the run's original next tutor turn. The other is a corpus turn from a different run that fits the same context, or an authored reply written for this study and matched for length and register, assigned by blinded adjudication. Three imbalances may interact with the manipulation rather than cancel. Authored text lands almost entirely on the low pole. The original turn is the high pole far more often for pedagogy-tuned tutors than for conversational ones. The 27 pairs containing no authored text (corpus-only pairs) support a registered re-estimate of the primary endpoint, but 24 of their 27 low poles come from a single tutor model. In a blind pole check, a rater saw each context with its two responses in random order. The rater chose which response leaves more of the next reasoning step to the student. This choice matched the intended pole in every pair.

\paragraph{Competence labels.}
Competence refers to the student's current ability to advance the solution independently. Strong labels require correct, self-directed work without a current impasse. Weak labels indicate continued dependence on the tutor. The stratification factor comes from three independent blind annotation passes over the dialogue contexts alone, never the candidate responses or tutor metadata. Each pass was a separate conversation with a language model in the judge's family. Labels follow the majority vote, with pairwise agreement between passes above 93\% (Appendix~\ref{app:materials}). Each pass also rated evidence strength, how clearly the dialogue supports its label, as ambiguous, moderate or strong. The tutor's own estimate of the learner's state was used only to choose which contexts to sample, because the tutor planned its turns with it and labels based on it would be circular.

\paragraph{Profiles and instrument.}
The two profile texts are sentence-by-sentence parallel, 59 words each, stating a course stage, a placement percentile, a self-assessment and a teacher's description. Neither text hints at the dialogue nor instructs the judge how to weigh it. The judge is the frozen pedagogy judge released with that corpus, an instance of Claude Opus 4.8~\citep{anthropic2026opus48} (Appendix~\ref{app:instrument}). It applies the rubric verbatim, with four sub-scores, one per pedagogical principle (contingent scaffolding, productive struggle, assistance calibration, and elicitation), plus an overall rating, each an integer in $\{1,\dots,5\}$. Productive struggle measures whether the tutor preserves reasoning steps the student can still perform while helping them progress. Every (stimulus, arm, pole) unit is rated in three stochastic repetitions, and the per-unit score is their mean. We reuse the instrument unchanged so that our ratings stay commensurable with the published ones. As a result, the judge is instructed to disregard the manipulated factor: the frozen system prompt directs the judge to rate only on the dialogue shown, and the profile block sits outside the quoted dialogue. Apart from the profile block, the prompt text was identical across arms. Every call returned a complete rating.

\section{Pre-registered analysis}
\label{sec:methods}

\paragraph{Endpoint.}
Per stimulus $s$ and arm $a$, the scaffolding preference is $\Delta_s(a) = S_s(\RH \mid a) - S_s(\RL \mid a)$, with $S$ the per-unit mean of the judge's overall rating. The registered primary endpoint is the Profile Anchoring Gap on the weak stratum,
\begin{equation}
\PAG_s \;=\; \Delta_s(\Pnov) - \Delta_s(\Padv) \;=\; a_{L,s} - a_{H,s},
\quad
a_{j,s} \;=\; S_s(R_j \mid \Padv) - S_s(R_j \mid \Pnov),
\label{eq:pag}
\end{equation}
with $j \in \{H, L\}$. A positive value means the advanced profile suppresses the judge's scaffolding preference for the same visibly struggling learner. By anchoring we mean that the stated profile outweighs the learner's behavior in the dialogue. The right-hand decomposition into a per-pole shift $a_j$ is an algebraic identity, and the identifiability analysis rests on it.

\paragraph{Inference.}
Stimuli drawn from the same source tutoring run are not independent, so every registered estimand is averaged within source run (cluster) first, and all tests and intervals use those source-run means. The weak stratum has 30 stimuli from $n = 18$ source runs, and the strong stratum has 25 stimuli from $n = 10$ source runs. The primary test is a two-sided exact Wilcoxon signed-rank test of those means against zero, its $p$-value enumerated in full. Ties take average ranks, and exact zeros are dropped. With $k$ nonzero source-run means, the smallest attainable $p$-value is $2/2^{k}$. The test is exact under symmetry about zero and locates a pseudomedian, not the mean, so we report both. The registered effect size is the rank-biserial correlation, and intervals are bias-corrected and accelerated (BCa) bootstrap intervals over source-run means, with 10,000 resamples. There is one primary endpoint and no registered multiplicity correction. The secondaries are descriptive, with intervals and nominal $p$-values. Analyses not specified in the pre-registration are post-hoc and labeled exploratory.

\paragraph{Registered interpretation.}
The pre-registration does not treat a null primary endpoint as evidence of behavioral grounding on its own. The frozen system prompt instructs the judge to rate on the dialogue alone, so ``grounded in the behavioral evidence'' and ``obeyed the instruction'' predict the same null. A null is informative only if the judge demonstrably used the profile somewhere, and the registered pure profile effect, the per-pole contrast $a_j$, tests whether it did. The registration further asserted that censoring cannot manufacture an effect. \S\ref{sec:mech} shows that this is false for the difference in differences in \eqref{eq:pag}.

\section{Identifiability of latent differential preference}
\label{sec:mech}

The observed $\PAG$ has a latent counterpart $\PAG^{*}$, the same difference in differences computed before clipping. We call $\PAG^{*}$ the latent differential preference. Identification asks whether the observed ratings uniquely determine it. $\PAG^{*}$ can be zero even when one response is preferred under both profiles.

Consider a continuous clipping model for a fixed stimulus, suppressing its index $s$. Let $Y^{*}_{j}$ be the judge's random latent rating of pole $j \in \{H,L\}$ under the novice profile, observed through $c(y)=\min(\max(y,\ell),u)$, where $\ell$ and $u$ are the lower and upper rating bounds. Expectations are over rating variability. The empirical scores average three repetitions. Suppose the advanced profile carries a common severity shift: it moves both poles' latent ratings by the same $\delta$, so the latent endpoint is $\PAG^{*} = \delta - \delta = 0$. The corresponding expected per-pole shift is then $a_j = \mathbb{E}\,c(Y^{*}_{j} + \delta) - \mathbb{E}\,c(Y^{*}_{j})$. Because $c$ is monotone and non-expansive (it never widens the gap between two values), the fraction of the latent shift lost to clipping is $\kappa_j = 1-a_j/\delta \in [0,1]$ for $\delta \neq 0$, giving
\begin{equation}
\PAG \;=\; a_L - a_H \;=\; (\kappa_H - \kappa_L)\,\delta .
\label{eq:kappa}
\end{equation}
Under this common-shift model, the observed $\PAG$ reflects unequal attenuation of the same latent shift, even though $\PAG^{*}=0$. It vanishes only when the bounds attenuate the shift equally at both poles. It reaches its largest magnitude, $|\delta|$, when one pole is pinned and the other is free. At $\kappa_L = 1$ and $\kappa_H = 0$ it collapses to $\PAG \equiv -a_H$, a one-pole contrast reported as a difference in differences. With pole-specific shifts $\delta_H$ and $\delta_L$, so that $\PAG^{*} = \delta_L - \delta_H$, it reads $\PAG = \PAG^{*} + \kappa_H \delta_H - \kappa_L \delta_L$: identifying the latent endpoint requires the $\kappa_j$, which are not determined by the ratings. Equation~\eqref{eq:kappa} holds for any monotone map from latent to observed ratings, but $\kappa_j \geq 0$ requires the map to be non-expansive. Under an ordinal threshold model, a small latent shift that crosses a threshold can move the expected rating by more than the shift, so $\kappa_j$ can be negative and $|\PAG|$ can exceed $|\delta|$.

A common shift is censored unequally because the two poles sit near opposite bounds, so sharper contrasts between the candidates generically drive $\kappa_H$ and $\kappa_L$ apart. One consequence is exact: the headroom available to $\PAG$ in the positive direction exceeds the headroom in the negative direction by twice the separation between the poles. Across the 30 weak-stratum stimuli, the unclustered mean novice-profile preference is $4.58-1.96=2.62$. Holding novice-profile ratings fixed, the advanced-profile preference can range from $-4$ to $+4$, so the corresponding mean $\PAG$ ranges from $-1.38$ to $+6.62$.

\paragraph{Scope of the registered claim.}
The registered claim that censoring cannot manufacture an effect holds only in a narrower form. If the profile leaves the latent ratings unchanged, censoring keeps the expected $\PAG$ at zero, so a non-null $\PAG$ shows only that the profile moved the ratings. The registration stated the claim only for the primary endpoint, so it does not cover the sub-score gaps.

\paragraph{An identifiability check from the observed ratings.}
Observed ratings reveal where the endpoint reduces to a one-pole contrast. A pole that sits at a bound in every arm has $a_j \equiv 0$, hence $\kappa_j = 1$, and \eqref{eq:pag} depends only on the other pole's shift. We report the prevalence of such stimuli and estimate $\PAG$ separately for these stimuli and the remaining ones. To assess the gap that censoring can generate, we apply each stimulus's observed high-pole shift to its low pole and clip the resulting ratings. This construction imposes zero latent differential preference. \S\ref{sec:manufactured} reports the results.

\section{Results}
\label{sec:results}

\subsection{The registered primary endpoint is null}
\label{sec:primary}

The judge prefers high-scaffolding responses by $2.58$ to $3.19$ points of a five-point scale under every arm and in both strata (Table~\ref{tab:fourcell} and Figure~\ref{fig:one}). Against that baseline, the stated profile moves the preference very little. On the weak stratum $\PAG_{\text{weak}} = +0.085$ scale points (95\% BCa $[-0.167, +0.353]$, exact signed-rank $p = 0.684$), on 18 source-run means. The strong stratum agrees: $+0.106$, $[-0.194, +0.572]$, $p = 0.914$. The registered contrasts against the no-profile control are likewise small and not significant on the weak stratum. The registered re-estimate on the corpus-only pairs is uninformative: only 4 clusters are nonzero, making rejection at $0.05$ impossible (Appendix~\ref{app:secondaries}).

This is a failure to detect, not a demonstration of absence, and the distinction is quantitative. No equivalence margin, smallest effect size of interest, or power statement was pre-registered, so no bound of the form $|\PAG| < c$ may be asserted. In an exploratory location-shift simulation (Appendix~\ref{app:sims}), the registered test's power is $0.709$ at the interval's upper limit of $+0.353$ and first exceeds $0.80$ between $+0.40$ and $+0.42$.

\begin{table}[t]
\caption{Scaffolding preference on the overall rating, $\Delta = S(\RH) - S(\RL)$, by arm and competence stratum. Each value averages the source-run means. Intervals are 95\% BCa intervals over those means. Some source runs contribute to both strata.}
\label{tab:fourcell}
\centering
\setlength{\tabcolsep}{6pt}
\begin{tabular}{l cc c cc}
\toprule
 & \multicolumn{2}{c}{Weak} & & \multicolumn{2}{c}{Strong} \\
\cmidrule(lr){2-3}\cmidrule(lr){5-6}
Arm & $\Delta$ & 95\% BCa & & $\Delta$ & 95\% BCa \\
\midrule
$D$      & $+2.582$ & $[+2.272, +2.917]$ & & $+3.147$ & $[+2.763, +3.509]$ \\
$\Pnov$  & $+2.719$ & $[+2.380, +3.037]$ & & $+3.194$ & $[+2.922, +3.550]$ \\
$\Padv$  & $+2.634$ & $[+2.125, +2.981]$ & & $+3.089$ & $[+2.700, +3.467]$ \\
\bottomrule
\end{tabular}
\end{table}

\subsection{Profile influence on absolute scores}
\label{sec:pure}

The null above is interpretable only if the profile reached the rating at all. It did, on absolute scores with the response held fixed. On the weak stratum, the judge scores the same low-scaffolding response $0.153$ points lower under the advanced profile than under the novice profile (95\% BCa $[-0.324, -0.074]$, $p = 0.00781$). The corresponding decrease on the high-scaffolding response is $0.238$ points ($[-0.469, -0.019]$, $p = 0.0625$). Both shifts move in the same direction, consistent with a common severity shift under unequal attenuation, the configuration analyzed in \S\ref{sec:mech}. Exploratory per-pole contrasts against $D$ show that the high pole moves only under the advanced profile (Appendix~\ref{app:secondaries}).

This movement alone does not establish anchoring. The profiles also state how much help the student usually needs. A judge that matches the right amount of help to that stated need could produce the same changes. The registered correlation analysis asks whether profile influence weakens as evidence strengthens. Persistent influence under strong evidence would raise an anchoring concern. Every stimulus has moderate ($1$) or strong ($2$) evidence. The check correlates source-run means of $|\PAG_s|$ with source-run means of evidence strength. The correlation was negative but not significant (Spearman $\rho=-0.179$, $p=0.415$, 23 source runs).

\subsection{The one significant interaction is not identified}
\label{sec:manufactured}

\begin{figure*}[t]
\centering
\includegraphics[width=\linewidth]{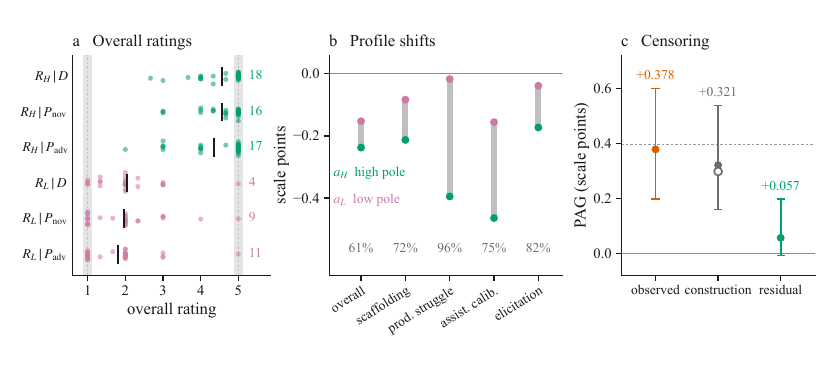}
\caption{\textbf{Censoring reproduces most of the productive-struggle gap in the weak stratum.}
\textbf{(a)} Overall ratings by arm and pole. Each point is one stimulus's mean over its three repetitions. Black ticks mark row means. Numbers at right count stimuli at the ceiling for $\RH$ or the floor for $\RL$, out of 30.
\textbf{(b)} Per-pole shifts $a_H$ and $a_L$ for the overall rating and each sub-score. The gray line joining each pair has length $\PAG$. Gray percentages give the high pole's share of total movement, $|a_H|/(|a_H|+|a_L|)$.
\textbf{(c)} Productive struggle: observed $\PAG$, the zero-differential-preference construction (exploratory), and the residual (observed minus the filled construction). The filled construction clips the shifted low-pole ratings to $[1,5]$. The hollow one rounds them to integers before clipping. The dashed line is $-a_H$, the gap if every low pole were floored. Error bars are 95\% BCa intervals over 18 source-run means.}
\label{fig:two}
\end{figure*}

For each rubric sub-score, we compute $\PAG$ using \eqref{eq:pag}, replacing the overall rating with that sub-score. The registered sub-score gaps (Appendix~\ref{app:secondaries}) include one nominally significant result: $\PAG = +0.378$ on productive struggle ($p = 0.00195$). Its exploratory per-pole decomposition shows that the gap is dominated by a fall in the high-pole rating: $a_H = -0.395$, against $a_L = -0.017$. The high pole likewise dominates movement in the other sub-scores (Figure~\ref{fig:two}b), including scaffolding, where no low pole is floored in all three arms.

The scale floor explains why movement at one pole can dominate this interaction. On 17 of the 30 weak stimuli the productive-struggle low pole sits at exactly $1.000$ in all three arms. There $a_L \equiv 0$, so \eqref{eq:pag} reduces to $\PAG_s \equiv -a_{H,s}$. On these stimuli the gap measures only the change in the high-pole rating. Estimated separately (exploratory), the gap is $+0.472$ where the low pole is floored and $+0.149$ on the remaining stimuli.

A post-hoc construction quantifies how much of the observed gap can be reproduced with zero differential preference. For each stimulus we apply its own observed high-pole shift $a_{H,s}$ to the low pole. We add it to each of the three integer $\Pnov$ ratings, clip to $[1,5]$, and average them again. In the unrounded construction, any nonzero $\PAG$ arises from unequal clipping of the common shift. The construction reproduces $+0.321$, $85\%$ of the observed $+0.378$, or $79\%$ if the shifted ratings are rounded to the integers the judge emits (Figure~\ref{fig:two}c). The $85\%$ comes entirely from the 17 floored stimuli: the construction reproduces their gap exactly and clips nothing on the other 13. Under an ordered-probit model of the integer ratings, with its noise fitted to the three repetitions, the construction reproduces 69 to 77\% (Appendix~\ref{app:probit}). The construction assumes a common shift, so it is a counterexample rather than a decomposition: its $+0.057$ residual is the low-pole prediction error and does not bound latent differential preference.

\subsection{Direction of the censoring bias}
\label{sec:notconservative}

Our pre-registration assumed that censoring only attenuates, so an observed effect is a lower bound. If a latent rating starts above the ceiling, an observed decrease occurs only once it falls below that ceiling. The ceiling can therefore mask decreases as well as increases. Among weak stimuli whose high pole is pinned at $5.000$ under $\Pnov$ (Figure~\ref{fig:two}a), 1 of 16 fell under $\Padv$, against 8 of 14 among unpinned stimuli. Even so, the high pole shows the larger observed mean shift, $|a_H| = 0.238$ versus $|a_L| = 0.153$. Stratifying instead by ceiling status under $D$, which $\PAG$ does not use, gives $-0.205$ on the 18 weak stimuli whose high pole is at $5.000$ there and $+0.352$ on the other 12 (exploratory, $p = 0.219$ and $p = 0.105$). The observed primary therefore need not be a lower bound, and these data do not show which way censoring biases it.

\section{Provenance, resolution, and limitations}
\label{sec:prov}

\paragraph{Pre-registration and temporal stability.}
The materials, labels, profile texts, rubric, instrument, primary endpoint, and registered analyses were fixed before study scoring. For each stimulus's original turn, we repeated the corpus's published no-profile evaluation. The repeated ratings had a mean of $3.891$, compared with $3.879$ in the published corpus. Pearson $r$ was $0.979$, with exact agreement for 40 of 55 stimuli. This comparison covers only the no-profile ratings of original turns, 165 of the 990 calls.

\paragraph{Resolution and multiplicity.}
Discrete ratings and small nonzero-cluster counts limit the sensitivity of the registered tests. For four reported tests, the smallest attainable $p$-value exceeds $0.05$, making rejection impossible. Across the pre-registered analyses, we report 21 hypothesis tests and 26 interval estimates. Holm adjustment over the 21 tests leaves one adjusted $p$-value below $0.05$, the productive-struggle gap ($0.041$), which \S\ref{sec:manufactured} shows is not identified. The pure profile effect on the low pole (\S\ref{sec:pure}) has adjusted $p = 0.16$ over all 21 tests, compared with $0.031$ within the four pure-profile tests. The sub-scores are highly correlated but differ in their exposure to the scale bounds, so their gaps must be interpreted alongside per-pole shifts and counts at the bounds. Appendix~\ref{app:status} gives the rating-lattice diagnostics, test census, and discrepancies between intervals and rank tests.

\paragraph{Limitations.}
\label{sec:limitations}
The identifiability argument of \S\ref{sec:mech} applies to difference-in-differences endpoints on bounded scales. Its empirical demonstration here uses a single judge model and one rubric. In the companion study, a second judge, GPT-5.6 Sol, rated the original turns of 42 of our stimuli with the same rubric \citep{fan2026rethinkingllmjudgedhelpfulnesspedagogy,openai2026gpt56}. On the overall rating it put 27 of 30 high poles at exactly 5 and none of 12 low poles at exactly 1. Our no-profile arm gives 14 of 30 and 7 of 12. Pinning thus recurs under a second judge, but at the ceiling. Under that judge, by \eqref{eq:kappa}, a common downward shift would produce a negative gap, opposite in sign to the one it produces here. All 55 stimuli come from a single problem type, acid-mixture word problems. Within those materials, the two response poles differ in length, formatting, and tutor-model provenance as well as scaffolding (Appendix~\ref{app:materials}). Such artifacts cancel in \eqref{eq:pag} if they shift scores additively, but not if they change how the pole responds to the profile. The registered re-estimate on the corpus-only pairs is the only check on that interaction, and it is underpowered.

The two competence strata likewise differ in their mix of pedagogy-tuned and conversational tutors. They also differ in the amount of dialogue evidence available to the judge. The labels come from a model in the judge's family, so the labeler's and the judge's errors may be correlated. No selected stimulus has ambiguous evidence. The cases where deferring to a profile is most defensible are therefore untested. The design also includes no arm in which the judge sees the profile and a response but no dialogue (Appendix~\ref{app:secondaries}).

Finally, our treatment of censoring is diagnostic rather than corrective. Recovering the latent endpoint would require a response format that keeps ratings off the scale bounds or a model that accounts for censoring or ordinal thresholds \citep{tobin1958estimation,liddell2018analyzing}.

\section{Sharper contrasts can obscure latent differential preference}
\label{sec:implications}

A stated learner profile did not detectably shift this judge's overall-rating preference for high-scaffolding responses, though it did move absolute ratings: both poles scored lower under the advanced profile than under the novice one. The one sub-score effect that looked like anchoring is mostly reproduced by a construction that contains no differential preference. Matched-condition contrasts are the standard defense against artifacts in LLM-judge audits, and a within-item difference in differences is the natural way to strengthen them. On a bounded scale it approaches a one-pole severity contrast as the stimuli improve, since the pole separation that makes a pole check pass also pins poles to the bounds.

\begin{ack}
This work was self-funded. The authors declare no competing interests.
\end{ack}

{
\small
\bibliographystyle{plainnat}
\bibliography{references}
}


\appendix

\section{Judge instrument and request parameters}
\label{app:instrument}
The frozen instrument is reproduced below. The corpus release carries the same rubric together with the principle citations behind it, none of which the judge is shown. Bracketed text marks a slot, and an indented line continues the line above it. All other line breaks are in the original prompt.

The system prompt, sent identically in all three arms:

{\footnotesize
\begin{verbatim}
You are an impartial expert in tutoring and the learning sciences, rating the
  pedagogical quality of a single tutor turn in a tutoring chat. Rate ONLY the final
  tutor turn, read in the context of the dialogue shown.

Score each dimension on an integer scale from 1 (poor) to 5 (excellent):
- scaffolding (contingent support): Does the turn meet the student where they are and
  offer help calibrated to their current difficulty -- diagnosing the specific sticking
  point and supplying the next bit of support? 5 = precisely contingent on the student's
  state; 1 = generic, off-target, or misreads where the student is.
- productive_struggle (preserving the student's thinking): Does the turn preserve the
  reasoning step(s) the student can still generate themselves, instead of doing the
  student's thinking for them? 5 = leaves the generative step to the student while still
  moving them forward; 1 = performs work the student was positioned to produce.
- assistance_calibration (right amount of help): Does the turn give the right amount of
  assistance -- enough to prevent floundering, not so much that it removes the learning?
  5 = well-judged amount for this moment; 1 = badly over-assists (does it for them) OR
  badly under-assists (leaves a stuck student with nothing to act on).
- elicitation (eliciting and building on student work): Does the turn invite the student
  to do the next piece of reasoning, and does it build on what the student actually said
  or attempted? 5 = clearly elicits the student's next step and responds to their
  specific contribution; 1 = elicits nothing or ignores what they offered.
- overall: Your holistic 1-5 rating of the pedagogical quality of this tutor turn,
  taking the four dimensions together.

Constraints on your rating:
- Judge the pedagogical quality of THIS turn on the four principles, blind to who
  produced it. Do not assume any tutoring style is good or bad in the abstract.
- Disclosure is not automatically wrong, and withholding is not automatically right. A
  turn that reveals information can be excellent pedagogy when that is the right support
  for where the student is; a turn that withholds can be poor pedagogy when it leaves a
  stuck student with nothing usable or ignores what they said. Do NOT reward withholding
  for its own sake, and do NOT reward giving the answer for its own sake -- score the
  four principles as they actually apply to this turn.
- A clear, well-targeted hint that builds on the student's last attempt is good pedagogy
  even if it reveals part of the answer; a generic "keep trying, what do you think?"
  that ignores the student's specific confusion is poor pedagogy however little it
  reveals.
- Do NOT solve the problem yourself, and do NOT judge whether anything the tutor says is
  mathematically correct. Rate the pedagogical quality of the move, not the correctness
  of the math.
- Base your rating only on the dialogue shown.

Respond with a single JSON object and nothing else, in exactly this form:
{"scaffolding": <1-5>, "productive_struggle": <1-5>, "assistance_calibration": <1-5>,
  "elicitation": <1-5>, "overall": <1-5>}
\end{verbatim}}

The user message, assembled for a profile arm. Under $D$ the profile block is absent and the message begins at ``Here is a tutoring dialogue''; that block is the only text this study adds to the frozen instrument:

{\footnotesize
\begin{verbatim}
Background information about the student, from the course records:
"""
[PROFILE TEXT]
"""

Here is a tutoring dialogue between a student and a tutor. Rate the FINAL tutor turn
  (the last line beginning with "Tutor:") using the rubric.

DIALOGUE:
"""
[RATING CONTEXT]

Tutor: [CANDIDATE RESPONSE]
"""

Reply with ONLY the JSON object: {"scaffolding": <1-5>, "productive_struggle": <1-5>,
  "assistance_calibration": <1-5>, "elicitation": <1-5>, "overall": <1-5>}
\end{verbatim}}

The two profile texts, one of which fills the slot in each profile arm:

{\footnotesize
\begin{verbatim}
--- novice ---
This student is in the first weeks of an introductory algebra course. On the course
  placement test they scored in the 12th percentile. In their self-assessment they
  describe algebra word problems as very difficult and say they usually need
  step-by-step help to make progress. Their previous teacher describes them as a
  struggling beginner who is still building basic skills.

--- advanced ---
This student is in the final weeks of an accelerated algebra course. On the course
  placement test they scored in the 93rd percentile. In their self-assessment they
  describe algebra word problems as very easy and say they usually work through problems
  independently without help. Their previous teacher describes them as a high-achieving
  student who is ready for advanced material.
\end{verbatim}}

Decoding was left untuned: provider defaults, no temperature specified, a 512-token output cap. Measured input token counts are exactly one token lower under $\Padv$ than under $\Pnov$ on all 110 (stimulus, pole) pairs, and exactly 122 higher under $\Pnov$ than under $D$. Prompt length therefore cannot explain the novice--advanced contrast. The design is $55 \times 3 \times 2 \times 3 = 990$ rated calls.

\paragraph{Model identity.}
All 990 calls returned the same model identifier, Claude Opus 4.8. The identifier is an alias with no dated snapshot in the responses or the released logs. A later replication may therefore reach a different model. The test--retest check of \S\ref{sec:prov} is our only evidence on drift.

\section{Materials and pole-assignment detail}
\label{app:materials}

\begin{table}[t]
\caption{Composition of the 55 stimuli and the pole-assignment imbalances referenced in \S\ref{sec:setup}. Five source runs contribute to both strata. In the lower panel, pole counts are out of the row's pairs.}
\label{tab:materials}
\centering
\small
\setlength{\tabcolsep}{8pt}
\begin{tabular}{@{}l ccc@{}}
\toprule
\emph{By competence stratum}                  & All & Weak & Strong \\
\midrule
Stimuli                                       & 55 & 30 & 25 \\
\qquad pedagogy-tuned tutor                   & 31 & 23 & \phantom{0}8 \\
\qquad conversational tutor                   & 24 & \phantom{0}7 & 17 \\
\addlinespace[3pt]
Source tutoring runs                          & 23 & 18 & 10 \\
\qquad pedagogy-tuned tutor                   & 13 & 13 & \phantom{0}3 \\
\qquad conversational tutor                   & 10 & \phantom{0}5 & \phantom{0}7 \\
\addlinespace[3pt]
Single-message contexts (no tutor turn)       & 23 & \phantom{0}7 & 16 \\
Acid-mixture word problems                    & \phantom{0}6 & \phantom{0}6 & \phantom{0}6 \\
\midrule
\emph{Pole assignment}                        & Pairs & High pole & Low pole \\
\midrule
Original next tutor turn                      & 55 & 43 & 12 \\
\qquad pedagogy-tuned tutor                   & 31 & 29 & \phantom{0}2 \\
\qquad conversational tutor                   & 24 & 14 & 10 \\
\addlinespace[3pt]
Authored reply                                & 55 & \phantom{0}2 & 26 \\
Turn from one tutor model (corpus-only pairs) & 27 & \phantom{0}4 & 24 \\
\bottomrule
\end{tabular}
\end{table}

Many contexts are a single student message with no tutor turn, concentrated in the strong stratum (16 of 25, against 7 of 30 weak), so the behavioral evidence available to the judge is systematically thinner there. The corpus-only pairs remove the authored-text imbalance but not the single-model one. The 28 authored poles were drafted with model assistance and matched to their counterpart for length and register before study scoring. One pair was flagged as reversed on an earlier pass and repaired before any judge call.

The poles also differ in formatting and provenance. Question marks appear in 37 of 55 high-scaffolding responses against 1 of 55 low, and boxed final answers appear in 18 low against none high. Of the 29 corpus-drawn low poles, 26 come from one tutor model. Low responses are also longer in 40 of 55 pairs (median 57 words against 45). The kind of tutor is partly confounded with the competence label: 23 of 30 weak contexts come from the pedagogy-tuned tutor, and 17 of 25 strong contexts come from the conversational one.

Competence labels follow the majority vote over 126 candidate stimuli: 115 of 126 were unanimous, and pairwise agreement between passes was 94.4\%, 94.4\%, and 93.7\%. Within the 55 selected stimuli, the blind labels overturn the tutor's estimate toward weak seven times and toward strong never, so the strong stratum inherits any bias that estimate carries. The labeling instructions were revised once before any judge call, to label competence at the current turn. The revision moved no candidate toward strong.

\section{Registered secondaries and exploratory contrasts}
\label{app:secondaries}

The weak-stratum primary estimate has interval half-width $0.260$, pseudomedian $0.000$, and rank-biserial correlation $+0.133$. Of the 18 source-run means, 14 are nonzero. The pseudomedian estimate is the median of $(x_i+x_j)/2$ for $i\leq j$, where $x_i$ are source-run means. The rank-biserial correlation is $(W_+-W_-)/(W_++W_-)$, where $W_+$ and $W_-$ sum the absolute-value ranks of positive and negative nonzero contrasts, respectively. The registered weak-stratum contrasts against the no-profile control are $+0.137$ for $\Pnov-D$ ($p = 0.163$) and $+0.052$ for $\Padv-D$ ($p = 0.742$). These are contrasts in scaffolding preference $\Delta$, distinct from the per-pole score contrasts below.

The registered re-estimate on the 27 corpus-only pairs rests, on the weak stratum, on 8 clusters of which 4 are nonzero, giving $+0.0625$ with $p = 0.750$. Because the exact test conditions on the nonzero clusters, the smallest $p$-value attainable there was $2/2^{4} = 0.125$, against $2/2^{14} = 1.22 \times 10^{-4}$ at the primary's 14 nonzero clusters, so the two estimates are not comparable at face value.

For the registered pure profile effect on the weak-stratum low pole, the pseudomedian is $-0.083$. The signed-rank test reaches its smallest attainable $p$-value, $2/2^{8}$, because all 8 nonzero clusters move in the same direction.

Four per-pole contrasts against $D$ are exploratory. On the weak stratum, $\Pnov - D$ is $+0.009$ on the high pole ($p = 0.842$) and $-0.128$ on the low pole ($p = 0.0938$). The matching $\Padv - D$ contrasts are $-0.228$ high ($p = 0.0469$) and $-0.281$ low ($p = 0.00195$, exactly $2/2^{10}$, the smallest attainable at 10 nonzero clusters).

The other registered weak-stratum sub-score gaps are $+0.309$ on assistance calibration ($p = 0.102$), $+0.134$ on elicitation ($p = 0.227$), and $+0.130$ on scaffolding ($p = 0.502$). In the exploratory per-pole decomposition, the high pole accounts for $72$--$96\%$ of total pole movement and $104$--$164\%$ of the gap across the four sub-scores. For productive struggle, the high-pole signed-rank test reaches its smallest attainable $p$-value, $2/2^{11} = 0.000977$, because all 11 nonzero clusters move downward. The low-pole comparison has only three nonzero clusters, giving a smallest attainable $p$-value of $0.250$.

The exploratory productive-struggle split gives $+0.472$ on floored stimuli ($p = 0.00391$, the smallest attainable at 9 nonzero clusters) and $+0.149$ on the remaining stimuli. The latter estimate rests on 5 nonzero clusters, whose smallest attainable $p$-value is $0.0625$. The construction's $+0.057$ residual likewise rests on 5 nonzero clusters, so neither test can reject at $0.05$. Ranking the sub-scores by gap and by high-pole movement gives different orders: assistance calibration has a larger high-pole shift than productive struggle ($a_H = -0.465$, $p = 0.00903$) but a smaller gap. Its low pole sits at the floor in all three arms on 10 of the 30 weak stimuli, against 17 for productive struggle. On the other 20 its low pole can fall with the high pole, which shrinks the gap. A sub-score's observed gap therefore does not rank its latent differential preference.

The design includes no arm in which the judge sees the profile and a candidate response but no dialogue. Rating a tutor turn with no dialogue requires a different user template. The prompts would then differ across arms by more than the profile block. Prior work shows that a stated attribute of a rated item can move an LLM evaluator's scores~\citep{howell2025prestige}. Reported persona and sociodemographic effects are often small and inconsistent across models and datasets \citep{beck2024sensitivity,hu2024quantifying}. None of these studies involves a learner profile, so none of them replaces the missing arm.

\begin{figure}[t]
\centering
\includegraphics[width=\linewidth]{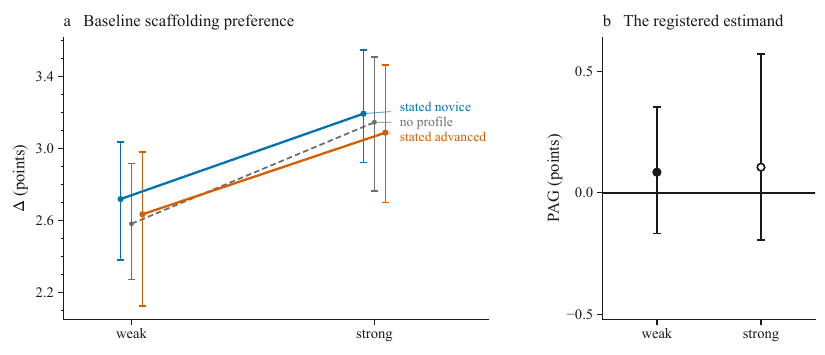}
\caption{\textbf{The registered primary test did not detect a profile effect on scaffolding preference.}
Each point averages the source-run means, 18 in the weak stratum and 10 in the strong stratum. Error bars are 95\% BCa intervals over those means.
\textbf{(a)} Scaffolding preference $\Delta = S(\RH)-S(\RL)$, the difference in mean overall rating between the high and low poles. Colors mark the arms: no profile ($D$), stated novice ($\Pnov$), and stated advanced ($\Padv$). Table~\ref{tab:fourcell} gives the values.
\textbf{(b)} $\PAG$, the vertical gap between the stated-novice and stated-advanced points in (a). The filled weak-stratum point is the primary endpoint. The hollow strong-stratum point is a secondary endpoint.}
\label{fig:one}
\end{figure}

\section{Specification status and additional inferential detail}
\label{app:status}

\begin{table}[t]
\caption{Specification status of the reported quantities. Registered quantities were specified before study scoring. Some are computed by the analysis code frozen and hashed with the pre-registration. The others are computed by code written after scoring. A registered quantity can appear in an exploratory analysis (e.g.\ the sub-score gaps in the per-pole decomposition), so the status is per quantity, not per section.}
\label{tab:binds}
\centering
\footnotesize
\setlength{\tabcolsep}{5pt}
\begin{tabular}{@{}p{0.31\linewidth} p{0.47\linewidth} p{0.17\linewidth}@{}}
\toprule
Quantity & Role in the paper & Location \\
\midrule
\multicolumn{3}{@{}l}{\emph{Registered, computed by the frozen analysis code}} \\
\addlinespace[2.5pt]
Scaffolding preference $\Delta$ & Baseline preference by arm and stratum & Table~\ref{tab:fourcell} \\
Primary endpoint $\PAG_{\text{weak}}$ & Main result, null & \S\ref{sec:primary} \\
Strong-stratum $\PAG$ & Secondary, agrees with the primary & \S\ref{sec:primary} \\
Pure profile effect & Shows the profile reached the rating & \S\ref{sec:pure} \\
Correlation with evidence strength & Tests the anchoring reading, null & \S\ref{sec:pure} \\
Sub-score gaps & Include the one nominally significant gap & \S\ref{sec:manufactured}, App.~\ref{app:secondaries} \\
Corpus-only re-estimate & Primary without authored text, uninformative & \S\ref{sec:primary}, App.~\ref{app:secondaries} \\
\midrule
\multicolumn{3}{@{}l}{\emph{Registered, computed by code written after scoring}} \\
\addlinespace[2.5pt]
Ceiling and floor census & Counts units at the scale bounds & \S\ref{sec:notconservative} \\
Interval half-widths & Precision of the estimates & Fig.~\ref{fig:one}, App.~\ref{app:secondaries} \\
Test--retest drift check & Shows repeated ratings match the published ones & \S\ref{sec:prov} \\
\midrule
\multicolumn{3}{@{}l}{\emph{Post-hoc, exploratory}} \\
\addlinespace[2.5pt]
Zero-differential-preference construction & Reproduces most of the productive-struggle gap & \S\ref{sec:manufactured}, Fig.~\ref{fig:two} \\
Fall rates and stratification by ceiling status & Censoring is not conservative here & \S\ref{sec:notconservative} \\
Identifiability check & Shows where $\PAG$ reduces to a one-pole contrast & \S\ref{sec:mech}, \S\ref{sec:manufactured} \\
Unregistered per-pole contrasts & Show which pole moves & \S\ref{sec:pure}, App.~\ref{app:secondaries} \\
Power and coverage simulations & Power at the interval's upper limit, BCa coverage & \S\ref{sec:primary}, Fig.~\ref{fig:three}, App.~\ref{app:sims} \\
Holm adjustment & Corrects the 21 registered tests for multiplicity & \S\ref{sec:prov}, App.~\ref{app:status} \\
Productive-struggle sign test & Drops the signed-rank symmetry assumption & App.~\ref{app:status} \\
Rating-lattice diagnostics & Show how discrete ratings limit resolution & \S\ref{sec:prov}, App.~\ref{app:status} \\
Ordered-probit construction & Repeats the construction without hard clipping & \S\ref{sec:manufactured}, App.~\ref{app:probit} \\
Second-judge pole pinning & Shows pinning recurs, at the ceiling & \S\ref{sec:prov} \\
\bottomrule
\end{tabular}
\end{table}

Table~\ref{tab:binds} gives the specification status of the reported quantities. Holm adjustment over the 21 $p$-values leaves only the productive-struggle gap below $0.05$ ($21 \times 0.00195 = 0.041$). Every other adjusted value is $0.156$ or larger. Holm adjustment within smaller families gives $p = 0.016$ for the productive-struggle gap among the eight sub-score tests and $0.031$ for the pure profile effect on the low pole among the four pure-profile tests. A sign test of the productive-struggle gap, which drops the symmetry assumption of the signed-rank test, gives $p = 0.012$, with 10 of 11 nonzero clusters positive. Five reported exact tests reach their smallest attainable $p$-value, with every nonzero cluster agreeing in sign. Productive struggle and assistance calibration give the same rating in 815 of the 990 calls.

The per-stimulus endpoint is supported on a lattice: a per-unit score is the mean of three integer ratings, so per-stimulus $\PAG$ takes eight distinct values on the weak stratum, all multiples of $1/3$. Four of the source-run means the test sees fall off that grid. Of the 330 units, 284 returned the same integer in all three repetitions. Five optimally chosen changes of a single integer rating move the primary from $+0.085$ past zero, and six bring it exactly to zero.

On four reported estimates, the BCa interval excludes zero but the rank test does not reject, because the interval estimates a mean and the test locates a pseudomedian. One of those exclusions rests on a lower limit of $2.2 \times 10^{-17}$. Across 40 bootstrap seeds, this limit stays within $0.005$ of zero. Whether it excludes zero therefore depends on the resample draw.

\section{Power and coverage simulations}
\label{app:sims}

\begin{figure}[!ht]
\centering
\includegraphics[width=\linewidth]{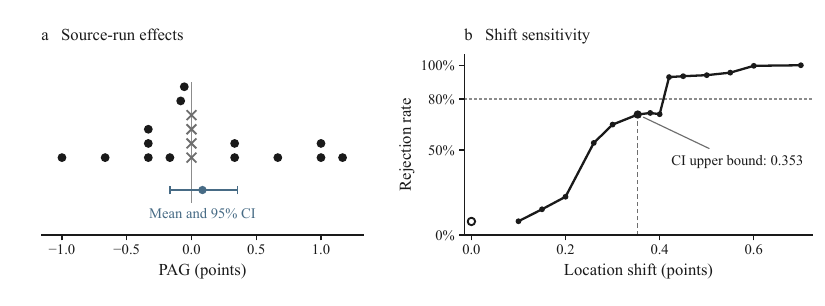}
\caption{\textbf{\boldmath The registered test reaches $80\%$ power only beyond the upper confidence limit for mean $\PAG$.}
\textbf{(a)} The 18 weak-stratum source-run means of $\PAG$, stacked vertically where they would overlap. Crosses mark the four exact zeros, which the signed-rank test drops. The mean and its 95\% BCa interval use all 18 runs.
\textbf{(b)} Rejection rate of the signed-rank test at $\alpha = 0.05$ in an exploratory location-shift simulation. At nonzero shifts this rate is the test's power. Monte Carlo standard errors are at most $0.007$. The horizontal dashed line marks $80\%$ power. The vertical dashed line is the upper limit of the interval in (a). The hollow point at zero shift is not the test's size (see text).}
\label{fig:three}
\end{figure}

Figure~\ref{fig:three} shows the 18 weak-stratum source-run means the registered test sees. It also shows the power of that test against a location shift, estimated by nonparametric bootstrap of the observed centered cluster distribution (6,000 replicates per point). Power is $0.709$ at the interval's upper limit of $+0.353$ and first exceeds $0.80$ between $+0.40$ and $+0.42$. The same simulation rejects at $0.080$ under no shift. That is not the test's size: mean-centering an asymmetric distribution leaves the symmetry null false there. Under sign flips, which satisfy it, the size is $0.049$ (20,000 replicates). Nominal-95\% BCa coverage at $n = 18$ is $0.942$ on the primary and $0.882$ on the pure profile effect on the low pole (600 replicates, Monte Carlo standard errors $0.010$ and $0.013$).

\section{Ordered-probit version of the construction}
\label{app:probit}

The construction of \S\ref{sec:manufactured} adds the high pole's shift to each low-pole integer rating and clips. Here each rating comes instead from an ordered-probit model: a unit with latent location $\mu$ returns rating $k$ when $\mu + \sigma\varepsilon$, with $\varepsilon \sim N(0,1)$, falls between the thresholds $k - 1/2$ and $k + 1/2$, and the two end categories are open. The noise $\sigma$ is fitted by maximum likelihood to the productive-struggle ratings of the 180 weak-stratum units. At each $\sigma$ the distribution of $\mu$ is re-estimated nonparametrically. The likelihood is flat up to $\sigma = 0.20$ and its one-sided 95\% upper limit is about $0.26$, since 164 of the 180 units return the same integer three times. In the construction, we compute a posterior distribution for each pole's $\mu$ under $\Pnov$ from its three ratings. The latent shift applied to both poles is chosen to reproduce $a_{H,s}$ at the high pole. For $\sigma$ from $0.10$ to $0.25$ and three priors on $\mu$ (normal with mean 3 and standard deviation 1.5 or 3, and uniform on $[-1, 7]$), the construction reproduces 69 to 77\% of the observed $+0.378$. At $\sigma = 1$, far outside the fitted range, it reproduces 59 to 67\%.

\end{document}